\documentclass[letterpaper, 10 pt, conference]{ieeeconf}  

\usepackage{graphicx}
\usepackage{amsmath}
\usepackage{amssymb}
\usepackage{booktabs}
\usepackage{url}
\usepackage{tikz}
\usetikzlibrary{arrows.meta, positioning, fit, backgrounds, calc,
  decorations.pathreplacing}

\IEEEoverridecommandlockouts                              

\title{\LARGE \bf
Draft: A Parametric Tool for Robot Design Exploration
}

\author{David Nguyen$^{1}$, Marcelo Coelho$^{2}$, and Sangbae Kim$^{1}$
\thanks{$^{1}$Department of Mechanical Engineering,
        Massachusetts Institute of Technology}%
\thanks{$^{2}$Department of Architecture,
        Massachusetts Institute of Technology}%
}

\begin{document}

\maketitle
\thispagestyle{empty}
\pagestyle{empty}

\begin{abstract}
Robot performance is often limited by the cost of iterating on morphology and
control together, since every computer-aided design (CAD) change
has to be carried into a simulation-ready model before control work begins.
Co-design methods attempt to close this gap, but each uses a model generator written for a single platform or lack the use of real-world data to suggest that designs are plausible. We present Draft, a parametric
generation tool whose generalized engine compiles any parametric tree of serial chains into a simulation-ready MJCF model, without CAD. It allows engineers to explore design tradeoffs through easily adjustable models and evaluate how changes influence controller performance. Draft grounds the free parameters of each design using trends fitted to a survey of $114$ actuators and $49$ published robot descriptions, so that a generated robot is anchored to real-world hardware. We validate those trends wholistically by building twins of four off-the-shelf robots, whose masses agree to $1.10\times$ geometric mean fold error. Finally, we demonstrate how Draft exposes design tradeoffs by evaluating three quadrupeds through a two-stage reinforcement learning curriculum.
\end{abstract}

\section{Introduction}

Integrating control and hardware design is slow because there is no middle layer
between computer-aided design (CAD) and the robot description files which a simulator
consumes. An engineer builds a CAD assembly, exports a
robot description, and hands it to a controls engineer, who builds a controller
and reports back. Each turn waits on a CAD export and a training run, so it
runs only a handful of times. The decisions that set a machine's ceiling, like leg length or
actuator placement, are therefore settled too early and with too little
information. The consequences surface later, in poor controller performance from a robot hitting its limits.

Co-design addresses this directly, optimizing design parameters and control together~\cite{spielberg2017functionalcooptimizationof, ha2018computationalcooptimizationof, dinev2022aversatilecodesign}.
The formulations vary by treating the two jointly or in an inner and outer loop. Despite co-design structure, the model generation is often specific to a single platform limiting its generalized use. Hardware constraints are often considered in model generation, but only loosely: motor-mass trends fitted to a handful of motors, and structure assigned a single density ~\cite{fadini2021computationaldesignof, howard2020diversitybaseddesignassist, dinev2022aversatilecodesign}.
The constraints an engineer meets when designing hardware are largely absent from these co-design strategies, providing little indication a robot model is possible to build.

\begin{figure}[t]
    \centering
    \includegraphics[width=\linewidth]{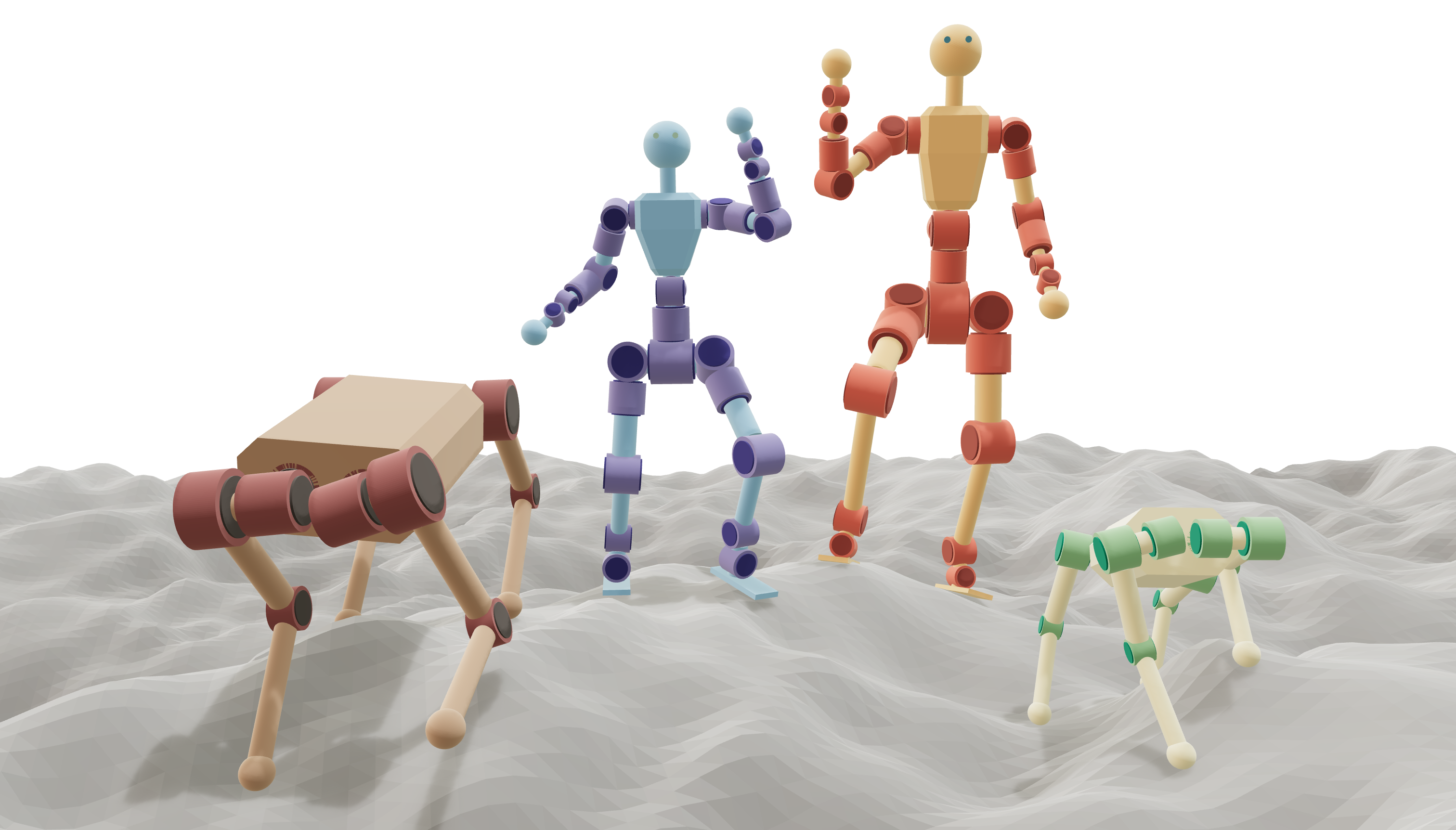}
    \caption{Draft compiles medium-fidelity robot models from a small set of specifications into a simulation-ready MJCF model. One parameter edit propagates through visual, collision, inertial, and actuation properties.}
    \label{fig:march}
\end{figure}

We present Draft, a parametric generation tool that enables rapid design
exploration using medium-fidelity robot models whose parameters are grounded in
real data (Fig.~\ref{fig:march}). With Draft, a robot is described by a
parameter file and a kinematic tree, which the generator compiles into a MuJoCo
MJCF~\cite{mujoco} model built from basic primitives and a lofted root mesh. The
CAD-to-simulation step disappears, so a family of designs that differ in gear
ratio, link length, and actuator choice is generated in milliseconds.

Our contributions are:

\begin{enumerate}
    \item A generalized MJCF generator that propagates every dimensional
    and actuator choice through visual, collision, and inertial properties by
    construction, with a humanoid and a quadruped as example models.
    \item A survey of $114$ integrated actuators and $49$ published robot
    descriptions, fitted into trends that ground a design's free parameters in reasonable values. The pipeline is validated end to end by rebuilding four off-the-shelf platforms.
    \item A demonstration that generated models hold measurable tradeoffs. Three quadrupeds with varying scale, stance, leg length, and actuator choice reach measurably different capabilities under a single learning environment.
\end{enumerate}

\section{Related Work}

\subsection{Co-Design Pipelines}

Parametric model generation is rarely studied independently; it is usually included in co-design pipelines, which shape its implementation. Model generation is tailored to the co-design optimizer rather than to an engineer who needs to convert the output to a manufacturable robot. Existing strategies each have their limitations.
Ghansah et al. insert design parameters as extra joints into a fixed kinematic
tree, so link lengths and inertias vary without rebuilding the model~\cite{ghansah2023humanoidrobotcodesign}. Including model parameters in symbolic dynamics functions integrates naturally with traditional optimal control but is less directly compatible with reinforcement learning~\cite{spielberg2017functionalcooptimizationof, ha2018computationalcooptimizationof, dinev2022aversatilecodesign}.
Vanteddu et al. drive a parametric CAD master model, which gives the highest fidelity available but is slow to regenerate, brittle under significant changes, and requires starting over for new robots~\cite{vanteddu2024fromcadto}. The base-mesh deformation of Xu et al. suits contact-rich manipulator design, but
addresses surface detail rather than the coarse structural changes that dominate early-stage design~\cite{xu2021anendtoenddifferentiable}. Gupta et al. and Yuan et al. both generate bodies from simple capsule and cylinder primitives but optimize morphology rather than a detailed robot~\cite{gupta2021embodiedintelligencevia, gupta2022metamorphlearninguniversal, yuan2022transformactlearninga}. RoboGrammar from Zhao et al. shares our modular premise through its use of link primitives, but does not provide a detailed evaluation of how feasible the generated designs are to build~\cite{zhao2020robogrammar}.

Draft aims to address these limitations by operating as a combined generator and evaluator, but not a full co-design loop. We exclude the design optimizer allowing engineers to perform the iteration themselves as to accommodate their complex design constraints. 

The choice of evaluation controller constrains the generator as much as the outer loop does. Trajectory optimization dominates because symbolic models integrate cleanly with optimal control, but it returns an open-loop plan rather than the closed-loop controller~\cite{dinev2022aversatilecodesign, schaff2022neuralapproachesto, fadini2024makinguseof}.
Reinforcement learning answers that limitation directly, since it returns a policy rather than a plan, therefore evaluating controllability alongside capability~\cite{schneider2025taskdrivencodesignof, zhu2026codesignofstructural}.
Its limiter is the difficulty of generating realistic learning-ready models at scale. Draft is designed to address this problem.

\subsection{Actuator and Structural Scaling Laws}

We fit custom parameter equations rather than adopt existing laws because of the uniqueness of robot actuators and how little their scaling is studied. Past work focuses on direct-drive motors or gearboxes alone and do not cover fully integrated modules which are commonplace in robotics. Dermitzakis et al. regress hobby servos and Rodriguez et al. seven decades of electric motors, but leave out the gearbox~\cite{dermitzakis2011scalinglawsin, rodriguez2026scalingofelectric}.
Saerens et al. study the transmission, but separate of motor laws which excludes inherent coupling between the two~\cite{saerens2019scalinglawsfor}. A motor drawn from one fit and a gearbox from another may not always add up to a
viable module. Honsinger's sizing relation
$\tau = 2\pi\sigma r^{2}\ell$ and the actuator design rules built on it turn geometry into torque for the motor alone, and again exclude the gearbox~\cite{honsinger1987sizingequationsfor, seok2015designprinciplesfor, wensing2017proprioceptiveactuatordesign}.
Co-design inherits these modeling limitations, reusing near-isometric laws fitted to a few frameless motors~\cite{fadini2021computationaldesignof, howard2020diversitybaseddesignassist}.

The structural trends of robots are even less studied. Oke et al. fit allometric laws across bipedal robots~\cite{oke2026allometricscalinglawsfor}, but a whole-robot law does not cover individual links, limiting the accuracy of robot designs outside of the fit distribution. We fit equations to each type of robot link and separate each link's mass into actuator and structural components to further improve model fidelity.

\section{Parametric Model Generation}
\label{sec:generation}

\subsection{Robot Specification}
\label{sec:spec}

In our formulation, a robot is defined by three files. The first is a parameter file that specifies values for link lengths, densities, and motor classes etc. The second is a kinematic tree whose fields are arithmetic expressions which correspond to the parametric file's values. The third describes the root body mesh. Because the tree holds expressions rather than numbers, it can be written so that fewer
parameters have to be specified. A limb's radius can be linked to the motor that drives it rather than manually specified.

The generator that takes in these files carries no body plan of its own. Humanoids, quadrupeds, and any other articulated robot share the same generation code, making new robot morphologies easy to compile and iterate upon. Furthermore, compilation is cheap enough to enable vast sweeps over design parameters. On a laptop CPU the $29$-joint humanoid compiles to a validated MJCF in $202$\,ms and a $12$-joint quadruped in just $50$\,ms.

We chose the output is an MJCF rather than a URDF, because it states actuator parameters such as joint armature and gear ratio directly and automates inertia calculations.

\begin{figure}[t]
    \centering
    \includegraphics[width=\linewidth]{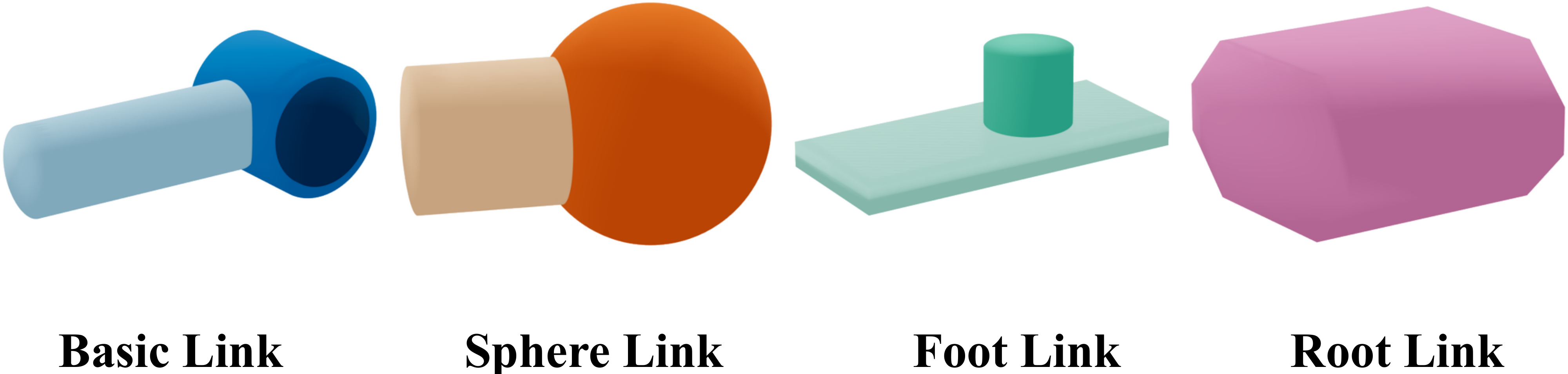}
    \caption{All generated Draft models are constructed from four main link primitives. The basic link includes a link and motor at its distal end, the sphere link terminates articulated chains with a contact point, the foot link does the same but with a support surface, and the root link holds custom geometry for limbs to extend from.}
    \label{fig:link_options}
\end{figure}

\subsection{Link Primitives and Actuation}
\label{sec:primitives}

Limbs are assembled from four link primitives
(Fig.~\ref{fig:link_options}), each with one job. The basic link is a structural
cylinder carrying at its far end the motor that drives the child joint, and
chaining it builds a serial articulated limb. The sphere link terminates a chain in a contact point. The foot plate terminates
one in a surface instead, so its contact sites span a support polygon. The root link allows the most geometric freedom for the placement of limbs and is therefore generated via a mesh. It is defined as cross-section polygons and planes which are lofted into a closed mesh. It stays tied to the same symbolic parameters, so a change in shoulder width propagates through to the mesh.

The inertial properties of the generated links which use primitives are calculated via MuJoCo's in-built \texttt{inertiafromgeom}. It computes a link's mass, and inertia tensor from the links geometry and an input density, saving the need to specify inertia directly in the generator. The root link's inertia values are input from \texttt{trimesh}~\cite{trimesh}. Link masses, motor classes, and root mesh density are fitted in Sec.~\ref{sec:survey}.

\label{sec:consistency}
Each joint is specified by a peak torque $\tau_\text{peak}$ and no-load speed $\omega_\mathrm{NL}$ which is modeled in the learning environment in Sec.~\ref{sec:comparison}. It drives each joint through a DC-motor model which constrains an actuator to a linear torque-speed curve. The joint's gains follow a hand-tuned scale, $k_p = 1.43\,\tau_{\text{peak}}$ and
$k_d = 0.071\,\tau_{\text{peak}}$, so a given tracking
error commands the same share of available torque on every joint.

The link library enables remotizing actuators by using fixed joint types, but does not support parallel mechanisms. For example, our quadruped's knee motor is placed at the hip, enabled by two combined modules which create a complex link without added degrees of freedom. 


\section{Fitting Model Parameters to Hardware}
\label{sec:survey}

\begin{figure}[t]
    \centering
    \includegraphics[width=0.9\linewidth]{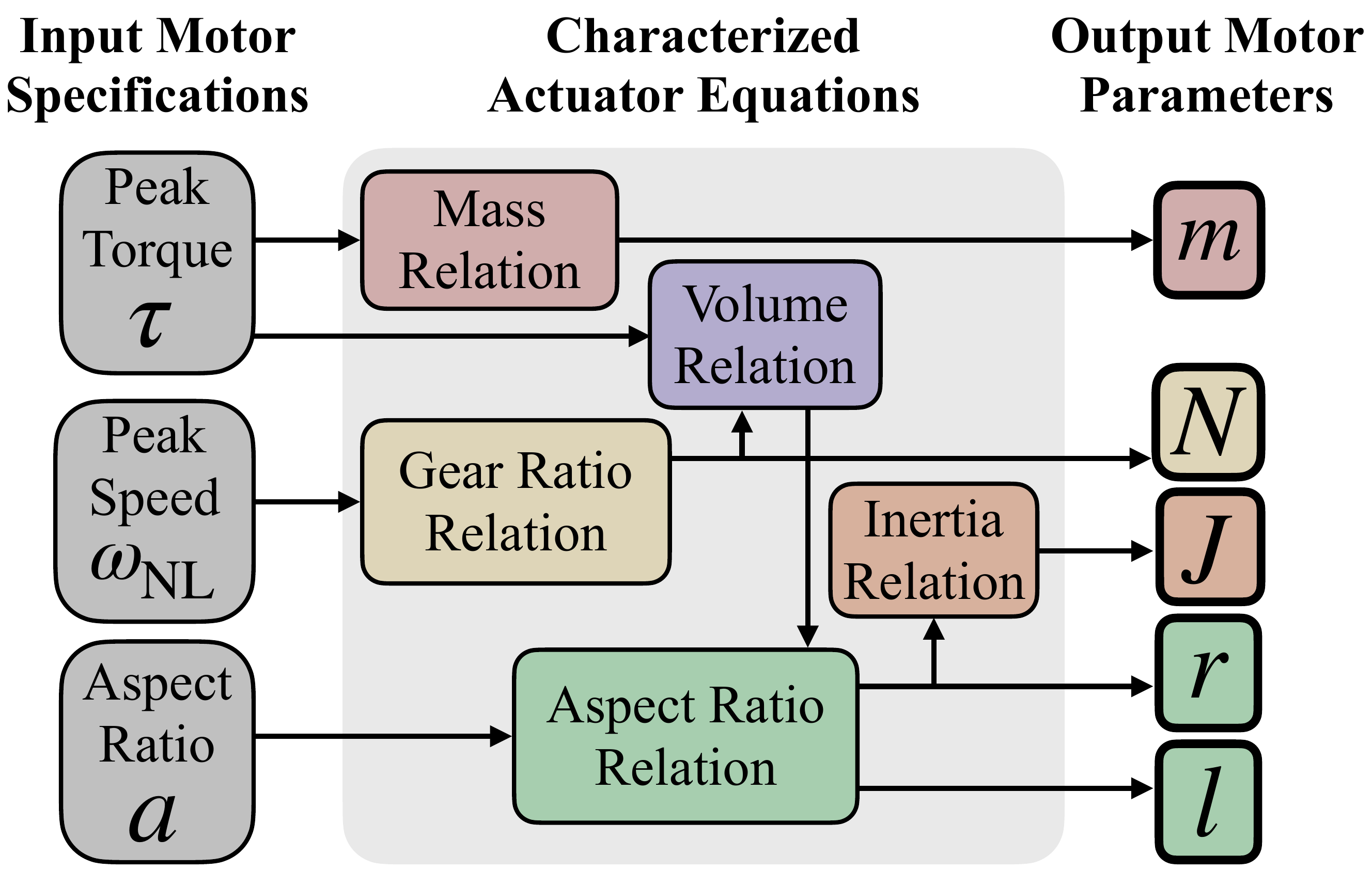}
    \caption{An actuator's parameters are derived from data-informed relations. The inputs are peak torque $\tau$, no-load speed $\omega_{\text{NL}}$, and aspect ratio $a$. Mass $m$ is estimated from $\tau$, and gear ratio $N$ from $\omega_{\text{NL}}$. Volume $V$ is estimated from $\tau$ and $N$; together with $a$, it determines radius $r$ and length $\ell$. Finally, the rotor inertia $J$ is derived from $r$.}
    \label{fig:actuator_gen}
\end{figure}

\begin{figure*}[t]
    \centering
    \includegraphics[width=\textwidth]{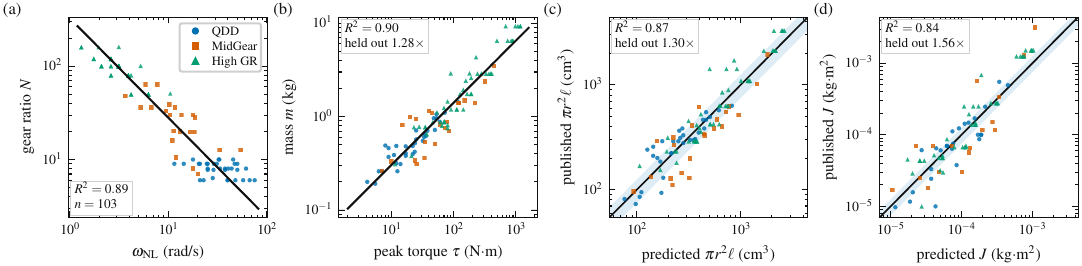}
    \caption{The four relations that dictate gear ratio, mass, volume, and rotor inertia plotted against the raw data. Panel (a) shows the power-law fit for gear ratio against no-load actuator speed, while (b) shows the fit for actuator mass against peak output torque. Panels (c) and (d) show prediction performance for volume and rotor inertia, respectively. Points are grouped by actuator class (QDD, MidGear, and High GR) and show agreement with the fitted predictions (black lines). The blue bands are $\pm25\%$ in (c) and (d).}
    \label{fig:laws}
\end{figure*}

The generator guarantees that a model is internally consistent, not that the design is plausible. Nothing in Sec.~\ref{sec:generation} stops a sweep from asking for an actuator with twice the torque density of anything on the market, and would report performance on impossible hardware. We therefore fit equations to surveys of real hardware and compile them back into the generation pipeline as guard rails. We perform two surveys, one on actuators and another on robot structure.

Every actuator fit is scored the same way, using leave-one-out (LOO) validation. We hold each entry
out in turn, refit on the rest, predict the held-out entry, and report the
geometric mean fold error over the $n$ entries,
\begin{equation}
    \mathrm{GMFE} = \exp\Big(\frac{1}{n}\sum_{i=1}^{n}
    \big|\ln(\hat y_i/y_i)\big|\Big),
    \label{eq:gmfe}
\end{equation}
where $\hat y_i$ is the held-out prediction and $y_i$ the published value, so
the GMFE is the typical factor by which a prediction misses in either
direction. A fold error of $1.10\times$ is a typical miss of $10\%$, measured on data excluded from the fit.

\subsection{Actuator Dataset}
\label{sec:motorsurvey}

We assembled $114$ rotary actuator entries, $111$ from $11$ different vendors
and three research actuators from publications. For each we record peak output torque $\tau_{\text{peak}}$,
no-load output speed $\omega_{\text{NL}}$, gear ratio $N$, module mass $m$, package
dimensions, and rotor inertia $J$. Every entry is an integrated module,
meaning motor, gearing, bearings, and housing in one package. We group them by transmission into \emph{QDD} (quasi-direct-drive,
reduction $6$ to $10$), \emph{MidGear} (integrated planetary, $7$ to $64$), and
\emph{High GR} (mostly strain-wave, $50$ to $161$). The catalog holds $33$ QDD,
$33$ MidGear, and $48$ High GR modules, spanning $2.5$ to $1180$\,N$\cdot$m and
$0.19$ to $9.3$\,kg.

The actuator population excludes high-performance rotor-stator pairs that are not integrated into a full actuator and therefore misses out on custom designs that beat our identified fits. Therefore our equations can be treated as a floor rather than a ceiling and act as a defensible baseline.


\subsection{Actuator Trends}
\label{sec:motorfits}

A module's mass and envelope carry the gearbox, housing, and bearings
alongside the rotor, so the trends borrow the form of bare-motor
theory~\cite{honsinger1987sizingequationsfor, reichert2009torquescalinglaws} where it
helps and let the catalog set the exponents. Fig.~\ref{fig:actuator_gen} provides a flow chart of how each value is derived from the resulting fits. The inputs for every generated motor are the peak torque $\tau_\mathrm{peak}$, no-load speed $\omega_\mathrm{NL}$ and aspect ratio $a$ which dictate all other motor parameters through five characterized relationships.

\subsubsection{Mass Relation}
Mass is fit to torque and its exponent is $0.666$, so mass grows
slower than torque contradicting linear scaling. That is steeper than either population of
the classical motor survey~\cite{dermitzakis2011scalinglawsin}, and the near-isometric hobby-BLDC relation ~\cite{fadini2021computationaldesignof}. Across this catalog, once output torque is fixed, the gear ratio that produces it barely influences mass suggesting near-equal contribution of gear ratio and motor mass to torque.

\subsubsection{Gear Ratio Relation}
Reduction is fitted against no-load speed over the geared catalog
(Tab.~\ref{tab:fitted}, Fig.~\ref{fig:laws}(a)), so naming a speed names the
gearbox that delivers it.



\begin{table}[t]
\centering
\caption{Fitted trends in pipeline order, with LOO fold error.}
\label{tab:fitted}
\small
\begin{tabular}{@{}llrrr@{}}
\toprule
Parameter & Fitted value & $n$ & $R^2$ & LOO \\
\midrule
Reduction $N$ & $327\,\omega_{\text{NL}}^{-1.066}$ & $103$ & $0.89$ & $1.33\times$ \\
Mass $m$      & $0.0653\,\tau^{0.666}$ & $103$ & $0.90$ & $1.28\times$ \\
Volume $V$    & $\tau^{0.700}N^{-0.167}/23948$ & $103$ & $0.87$ & $1.30\times$ \\
Radius $r$    & $(V/2\pi a)^{1/3}$ & $103$ & --- & $1.12\times$ \\
Length $\ell$ & $2ar$ & $103$ & --- & $1.14\times$ \\
Inertia $J$   & $27.2\,r^4$ & $88$ & $0.84$ & $1.56\times$ \\
\bottomrule
\end{tabular}
\end{table}

\subsubsection{Volume Relation}
Motor theory is used to choose which variables are related to the actuator volume. Airgap shear stress $\sigma$ over a rotor surface $2\pi r_a \ell$ at moment arm $r_a$ gives motor torque $\tau_m = 2\pi\sigma r_a^2\ell$, which is proportional to the volume enclosed by the airgap cylinder $V \propto \tau_m$~\cite{honsinger1987sizingequationsfor}. A full actuator's torque is $\tau = \tau_mN$ so we therefore expect $V \propto \tau N^{-1}$. A full actuator includes far more volume than just enclosed airgap volume so we leave both exponents free for the fit to identify. Volume goes as
$\tau^{0.70}N^{-0.17}$ (Tab.~\ref{tab:fitted}), weaker in each than the
relation of a bare rotor.

\subsubsection{Aspect Ratio Relation}
With volume specified from the torque and gear ratio, how that volume is distributed is left free. Therefore aspect ratio specifies a relationship between the actuator radius and length $a = \ell/2r$. 
Motor theory supports this since two rotors of equal $r^2\ell$ make the same air gap torque. Over the $103$ geared modules, radius and length computed from that aspect ratio and the volume above hold out to $1.12\times$ and $1.14\times$, against $1.19\times$ for both when geometry is instead taken from the mass trend.

\subsubsection{Rotor Inertia Relation}
Rotor inertia is how a gear ratio reaches the dynamics and is proportional to $r_a^4$ from dynamics principles so we use that relation and fit to the data. Over the $88$ modules that publish rotor inertia we get an exponent of $4.02$ on radius, close enough that we impose $4$ and fit only the constant Fig.~\ref{fig:laws}(d)). We reflect it through the
reduction and write $N^2 J$ as the joint's armature, which MuJoCo adds to the
diagonal of the joint-space inertia matrix.

\subsubsection{Actuator Fit Accuracy}
On held-out entries, reduction is predicted from no-load speed with a fold error of $1.33\times$, volume from
torque and reduction to $1.30\times$, and mass from torque to $1.28\times$
(Tab.~\ref{tab:fitted}). Because product siblings can share a rotor stator pair, holding out a single unit may only test interpolation within a vendor's own lineup. We test this by holding out a entire vendors to see if performance sizably degrades, which it does not with a similar fold error of $1.31\times$. The inertia trend is the loosest of the four, at $1.56\times$ but explainable. We size the rotor from the outer housing radius, and no vendor records whether a module is an inrunner or an outrunner which impacts the ratio of the rotor and housing radius. The fitted exponent still matches rigid-body physics through that noise, so every joint's armature $N^2 J$ scales expectedly.

\begin{figure*}[t]
    \centering
    \includegraphics[width=\textwidth]{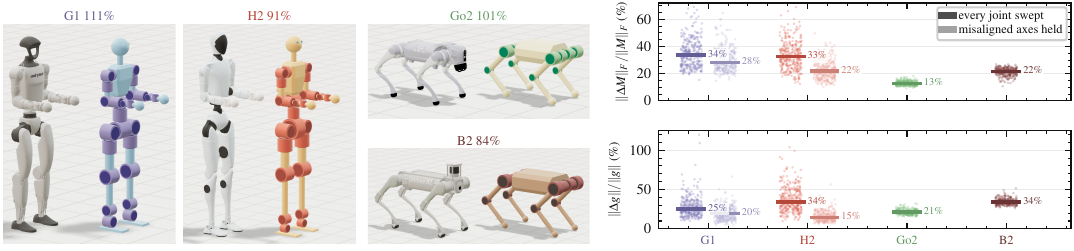}
    \caption{The generation pipeline run end to end against four off-the-shelf robots. The
    twin takes the vendor's geometry and declared torque and speed, and derives all other parameters. On the left are renders of the vendor and twin models with the twin's mass ratio as a percentage above. On the right is the dynamics over $300$ configurations per pair with inertia above and gravity below. The second spread for humanoids only holds the misaligned axes still.}
    \label{fig:twins}
\end{figure*}

\subsection{Structural Dataset}
\label{sec:structsurvey}

\begin{table}[t]
\centering
\caption{Structural mass by segment class, fit as $m = cL^{e}$ with $L$ in
meters, and its in-sample fold error.}
\label{tab:density}
\small
\setlength{\tabcolsep}{3pt}
\begin{tabular}{@{}l rrrr@{\hspace{0.8em}} rrrr@{}}
\toprule
& \multicolumn{4}{c@{\hspace{0.8em}}}{\emph{Humanoid}}
& \multicolumn{4}{c}{\emph{Quadruped}} \\
\cmidrule(r{0.6em}){2-5}\cmidrule{6-9}
Class & $n$ & $c$ & $e$ & Fold & $n$ & $c$ & $e$ & Fold \\
\midrule
Thigh         & $54$ & $13.8$  & $1.36$  & $1.38$ & $60$ & $28.3$ & $2.69$ & $1.43$ \\
Shank         & $50$ & $4.11$  & $1.68$  & $2.09$ & $60$ & $9.25$ & $2.38$ & $1.21$ \\
Hip link      & $56$ & $2.58$  & $1.05$  & $3.18$ & $72$ & $0.60$ & $0.51$ & $2.88$ \\
Head          & $16$ & $2.50$  & $0.63$  & $1.74$ & --- & --- & --- & --- \\
Upper arm     & $30$ & $9.40$  & $1.83$  & $1.88$ & --- & --- & --- & --- \\
Forearm       & $34$ & $0.22$  & $0^{*}$ & $2.88$ & --- & --- & --- & --- \\
Hand          & $28$ & $0.62$  & $0.33$  & $2.05$ & --- & --- & --- & --- \\
Foot          & $42$ & $0.19$  & $0^{*}$ & $3.00$ & --- & --- & --- & --- \\
Shoulder link & $44$ & $1.18$  & $0.76$  & $2.64$ & --- & --- & --- & --- \\
Wrist link    & $30$ & $2.33$  & $0.92$  & $1.42$ & --- & --- & --- & --- \\
Waist link    & $5$  & $0.69$  & $0^{*}$ & $1.57$ & --- & --- & --- & --- \\
\bottomrule
\end{tabular}
\end{table}

We parse $118$ published humanoid, biped, and quadruped descriptions from the
\texttt{robot\_descriptions} catalog~\cite{robotdescriptions} and from vendors'
own repositories, and keep $29$ humanoids and $20$ quadrupeds, spanning
$8.8$--$85.0$\,kg and $1037$ segments. We exclude models that depart far from our tree structure and morphology. 

This data requires pre-processing before fitting since robot description links contain the actuators that drive them. We desire only the structural parameters and therefore must remove the actuator mass from each description's links before evaluation. We do so by taking the torque and speed limits of each joint and approximating the motor's mass via Section \ref{sec:motorfits}'s relations. We then subtract it from the link's total mass, prepping it for fitting. We exclude entrees where the adjusted mass is negative, which is common for short links that have little structure in comparison to the motor. A limitation of this approach is the assignment of which link each actuator is attached to. We assume it follows our model convention for humanoids since it is difficult to parse from the models themselves. The quadrupeds are more consistent with the knee often remotized to the hip which we account for in our actuator mass removal.

\subsubsection{Structural Mass Trends}
The adjusted mass for each robot link is fitted per class against the segment's own length, $m = cL^{e}$ (Tab.~\ref{tab:density}), with $e$ floored at zero ($^{*}$) so that no member lightens as it lengthens. Humanoids and quadrupeds are fitted separately, since the humanoid trend predicts a quadruped thigh at over three times what it weighs. Robot torso's are fit using volumetric density rather than length and arrive at $716$\,kg/m$^3$ for humanoids and $654$\,kg/m$^3$ for quadrupeds.

\subsubsection{Structure Fit Accuracy}
Holding out each robot, we predict its total mass using the trends from Tab.~\ref{tab:density} and compare it with the mass in its official description. We do so by estimating every limb using its class's trend, the trunk by density and volume, and by adding the actuator mass that the joint torque implies to each link. We reach a geometric mean fold error of
$1.22\times$ over the $29$ humanoids, and $1.18\times$ over the $20$ quadrupeds. A baseline that compares the median robot mass with the held-out robot reaches a fold error of $1.42\times$ for humanoids and $1.77\times$ for quadrupeds. Fitting the two body plans separately is also validated, since the humanoid trends
applied to quadrupeds hold out at $1.38\times$.

The spread for the link relations is far wider than the actuator's for three main reasons. First, is our aforementioned assumptions in actuator link attribution. Second is the diversity in design choices. Designers can use a myriad of materials to design these robots adding variability. Third is our fold error from the actuator trends which propagates here from our actuator mass removal. Altogether, this leads to very high fold errors, but that is of little consequence in our later validation. In Sections \ref{sec:validator} and \ref{sec:comparison} we show that these trends can still reproduce off-the-shelf robots to moderate error and that the residual does not influence the high-level trends the tool is made to identify.

\subsection{Design Filter}
\label{sec:validator}

The fits of both surveys are compiled back into a filter that scores every generated model. It refuses actuators beyond its transmission family's extreme and the catalog's extreme. Aspect ratios outside of the fit range of $[0.28, 1.69]$ show up as warnings along with links over twice their classes structural density. For the full robot, warnings come from actuator mass fraction: median of $54\%$ for humanoids and $45\%$ for quadrupeds.


%

\subsection{Validation on Commercial Robots}
\label{sec:twins}

\begin{table}[t]
\centering
\caption{Twin error in $M(q)$ and $g(q)$ over $300$ poses, median and
variance, with misaligned axes held.}
\label{tab:twindyn}
\small
\setlength{\tabcolsep}{3pt}
\begin{tabular}{@{}l rrrr@{\hspace{0.8em}} rrrr@{}}
\toprule
& \multicolumn{4}{c@{\hspace{0.8em}}}{\emph{Inertia}}
& \multicolumn{4}{c}{\emph{Gravity}} \\
\cmidrule(r{0.6em}){2-5}\cmidrule{6-9}
& \multicolumn{2}{c}{Swept} & \multicolumn{2}{c@{\hspace{0.8em}}}{Axes Held}
& \multicolumn{2}{c}{Swept} & \multicolumn{2}{c}{Axes Held} \\
Robot & Med. & Var. & Med. & Var. & Med. & Var. & Med. & Var. \\
& (\%) & (\%$^2$) & (\%) & (\%$^2$) & (\%) & (\%$^2$) & (\%) & (\%$^2$) \\
\midrule
G1  & $33.9$ & $106$ & $28.3$ & $72.9$ & $25.1$ & $173$  & $19.5$ & $73.6$ \\
H2  & $32.6$ & $103$ & $21.9$ & $47.0$ & $34.1$ & $249$  & $14.5$ & $48.2$ \\
Go2 & $12.8$ & $2.4$ & ---    & ---    & $21.4$ & $8.4$  & ---    & --- \\
B2  & $21.7$ & $7.7$ & ---    & ---    & $34.1$ & $14.6$ & ---    & --- \\
\bottomrule
\end{tabular}
\end{table}

So far, all validation has been local and none evaluate how well the relations comose into a full robot. We therefore rebuild four off-the-shelf platforms, Unitree G1 and H2 ($33$ and $76$\,kg) and Go2 and
B2 ($16$ and $75$\,kg) using our generator's fit relations. All four are manufactured by Unitree and are platforms academic work often utilizes. We pass the generator the vendor's geometry and joint capabilities. No mass, density, inertia, or armature crosses, so every mass in the twin is derived (Fig.~\ref{fig:twins}). 

All four land inside $\pm25\%$ of their shipped mass, at $111\%$ (G1), $91\%$
(H2), $101\%$ (Go2), and $84\%$ (B2), for a geometric mean fold error of $1.10\times$.


Total mass agreement does not consider how it is distributed and namely if the twin's dynamics are at all similar. What influences a controller's performance most is the joint-space mass matrix $M(q)$ and gravity vector $g(q)$ of the robot's dynamics. Therefore, we compare both between the twin and target robots at $300$ configurations. The right panels of Fig.~\ref{fig:twins} plot the gravity and inertia Frobenius error for each configuration as a percentage. The rotor term is disabled in these comparisons, since only G1 publishes a per-joint armature. The median and variance of the inertia and gravity errors are presented in Tab.~\ref{tab:twindyn} with Go2 winning at $12.8\%$ and $21\%$.

Both quadrupeds have sizably lower median errors and variance compared to the humanoids which can be attributed to a mismatch in joint axes. G1's twin has $6$ joints with up to $16^\circ$ of misalignment while H2's had four up to $30^\circ$. Holding those joints still provides the second spread in Fig.~\ref{fig:twins} which bumps the medians down by $6$ to $19\%$ and the variance by $32$ to $200\%^2$. The exclusion of the misaligned axes pushes the dynamics disagreement for the humanoids closer to the same range as the quadrupeds while the rest can be attributed to the compounding error of more degrees of freedom.

\section{Evaluating Designs with Reinforcement Learning}
\label{sec:comparison}

\begin{figure}[t]
    \centering
    \includegraphics[width=\linewidth]{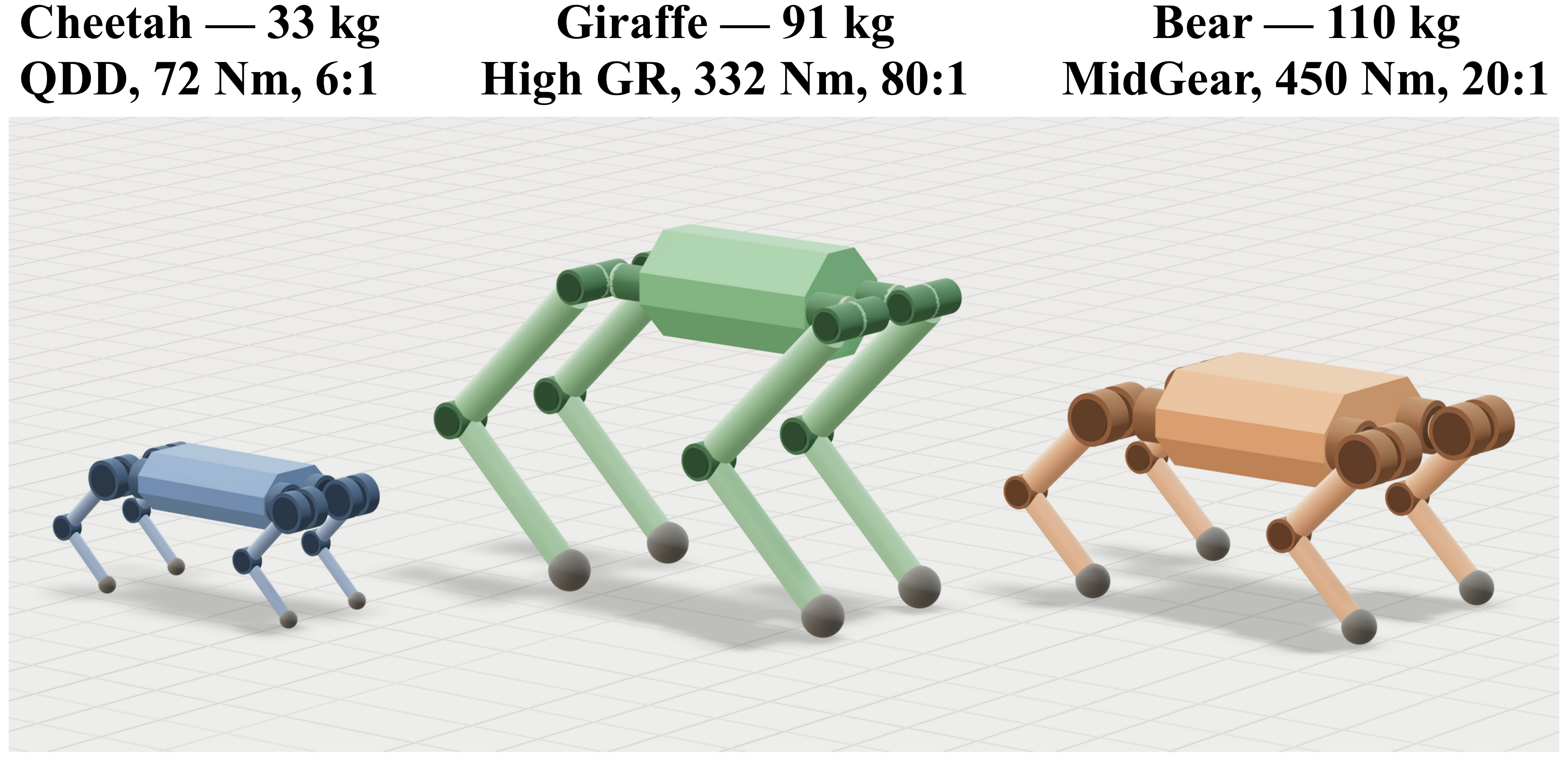}
    \caption{Three quadrupeds from one specification, differing in scale, leg
    length, and actuators, rendered in one scene at true relative size.}
    \label{fig:quadrupeds}
\end{figure}

Our model generation scheme is only useful if it shows how varying parameters influences controller performance. We demonstrate this, by generating quadrupeds with three distinct parameter sets ($\boldsymbol{\theta}_c, \boldsymbol{\theta}_g, \boldsymbol{\theta}_b$) and training them under one curriculum. To validate the robustness of the policies, we also perturb the inertial parameters post-training and analyze the results.

\subsection{Quadruped Designs}

Each quadruped design in Fig.~\ref{fig:quadrupeds} varies the levers a designer commits to early:
scale, stance, leg length, and motor choice. \textbf{Cheetah} is light and fast on QDD units at $6{:}1$. \textbf{Bear} is heavy and strong on mid-gear units at $20{:}1$, with a wide stance and over six times the cheetah's joint torque. \textbf{Giraffe} stretches the length axis, with legs more than twice the cheetah's on high-reduction actuators at $80{:}1$. All pass Sec.~\ref{sec:validator}'s design filter as feasible.


\subsection{Training Setup}

\begin{figure}[t]
    \centering
    \includegraphics[width=\linewidth]{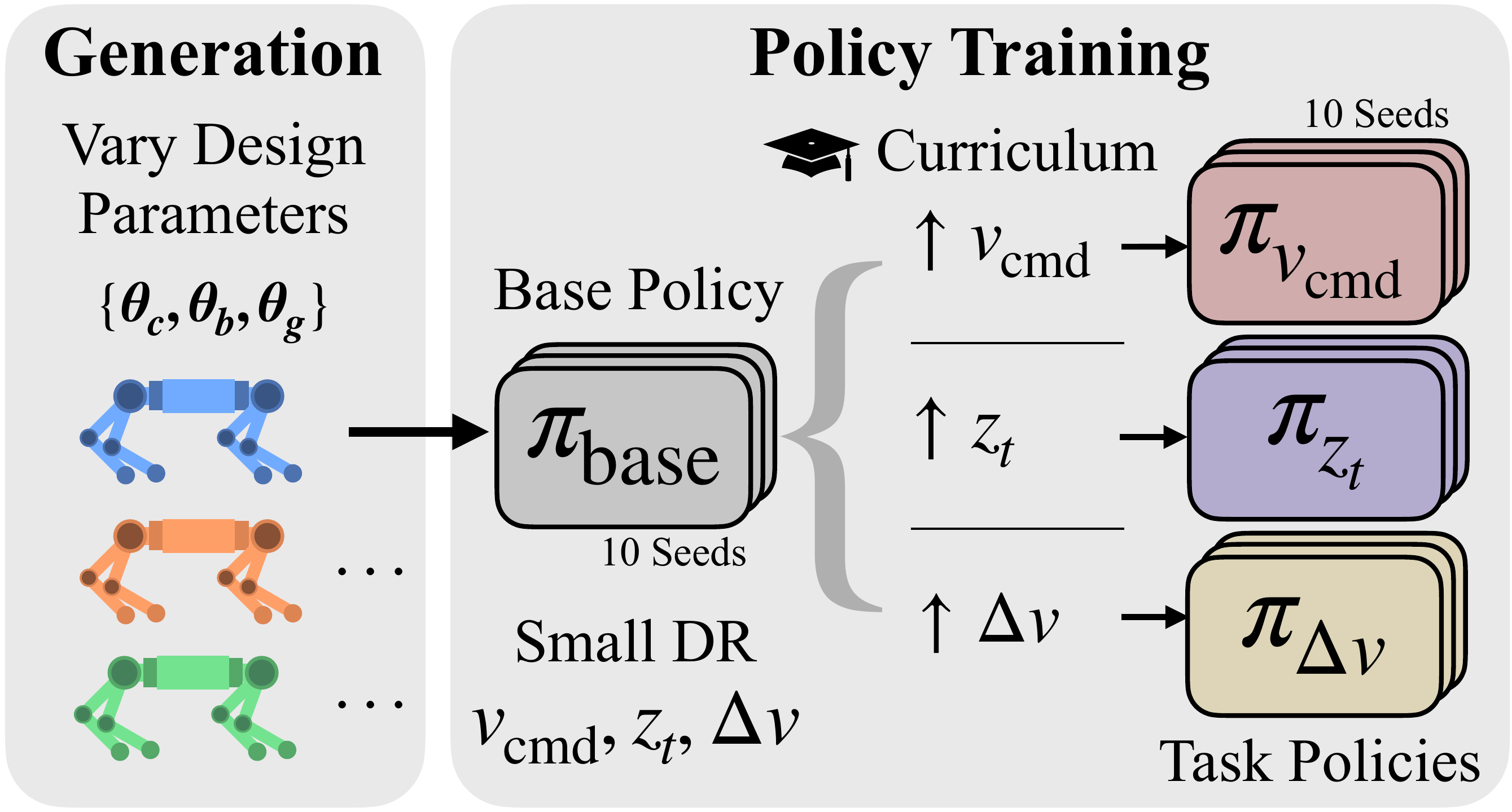}
    \caption{The same curriculum is used for each of the three robots. A base policy is trained over a small
    ranges of commanded speeds $v_{\text{cmd}}$, terrain heights $z_t$, and pushes
    $\Delta v$. Next, three task policies train the base policy further, each ramping one of axis. Every
    policy is trained from ten seeds.}
    \label{fig:training}
\end{figure}

Each design is trained end to end with proximal policy optimization~\cite{ppo}
in mjlab~\cite{mjlab}, ten times from independent seeds. The generated MJCF loads unmodified,
and every joint is driven through a DC-motor torque--speed curve from the
design's motor limits. Fig.~\ref{fig:training} provides an outline of how the robots are trained and evaluated through a two-stage learning curriculum.

\subsubsection{Base Policy Training}
Stage one trains a base locomotion policy on gently uneven ground $z_t$ under a
velocity-tracking objective, with intermittent pushes $\Delta v$ as domain randomization.
The policy is rewarded for tracking the commanded velocity $v_\mathrm{cmd}$, keeping its trunk
upright relative to the terrain, maintaining foot air time, and holding its base height. It
is penalized for foot clearance error, vertical body velocity, joint position
limits, upper-leg contact, and action rate.

The large range of quadruped sizes require us to scale the rewards accordingly since a stride natural for a
short leg is a shuffle for a longer one. Fixed reward terms bias the results towards specific robots, so instead, we scale the task by Froude similarity~\cite{alexander1983adynamicsimilarity}. For speeds and time we scale with $\sqrt{L}$ since legged bodies move similarly when $v^2/gL$ agrees while lengths are scaled by $L$. We define the quadruped scale as $s = L/0.48$\,m and utilize it in the reward function shape and weighting. Tab.~\ref{tab:rewards} shows these scaling factors and how they are used.

\begin{figure*}[t]
    \centering
    \includegraphics[width=\textwidth]{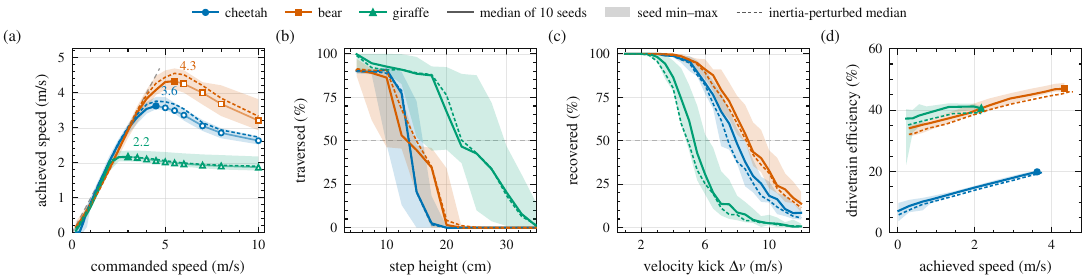}
    \caption{Four capability axes for the designs of Fig.~\ref{fig:quadrupeds},
    in absolute units. Lines are medians over ten training seeds and bands span
    the seed minimum to maximum; dashed lines are the same policies on the
    builds carrying the inertia error of Sec.~\ref{sec:twins}. (a) Commanded
    against achieved speed, hollow past each design's peak. (b) Traversal,
    ascent and descent pooled. (c) Recovery from a velocity kick. (d) Drivetrain
    efficiency up to each design's peak.}
    \label{fig:capability}
\end{figure*}

\begin{table}[t]
\centering
\caption{Stage-one reward terms; $s$ enters only where shown.}
\label{tab:rewards}
\small
\begin{tabular}{@{}l l l@{}}
\toprule
Term & Weight & Shaping \\
\midrule
Velocity tracking & $1.0$ & $\sigma = 0.5\sqrt{s}$\,m/s \\
Yaw-rate tracking & $1.0$ & $\sigma = 0.5/\sqrt{s}$\,rad/s \\
Upright           & $1.0$ & $\sigma = 0.45/\sqrt{s}$ \\
Air time          & $0.4$ & $t_\mathrm{max} = 0.35\sqrt{s}$\,s \\
Base height       & $0.4$ & $h^{*} = 0.43s$, $\sigma = 0.11s$\,m \\
Foot clearance    & $-2.0\,s^{-3/2}$ & $h_f^{*} = 0.10s$\,m \\
Vertical velocity & $-2.0\,s^{-1}$ & --- \\
Joint limits      & $-1.0$ & --- \\
Upper-leg contact & $-1.0$ & --- \\
Action rate       & $-0.05$ & --- \\
\bottomrule
\end{tabular}
\end{table}

\subsubsection{Task-Based Curricula}
Stage two takes the base policy and continues training using a curriculum meant to extract peak performance on one of three axes: peak velocity, terrain traversal, and stability to pushes. One of three environment variables is ramped for each curriculum: commanded velocity $v_\mathrm{cmd}$, terrain height $z_t$, and push magnitude $\Delta v$. For each, the other environment variables or targets are kept at the same value as the base policy environment. 
The three curricula are identical across the designs, such that the robot's themselves are evaluated rather than the training environment.


\begin{table}[t]
\centering
\caption{Capability, absolute then per leg length $L$; $\hat v = v/\sqrt{gL}$.
Medians over ten seeds; bold is best.}
\label{tab:capability}
\small
\setlength{\tabcolsep}{3pt}
\begin{tabular}{@{}l rr rr rr rr r@{}}
\toprule
& & & \multicolumn{2}{c}{Speed} & \multicolumn{2}{c}{Step}
& \multicolumn{2}{c}{Kick} & \\
\cmidrule(lr){4-5}\cmidrule(lr){6-7}\cmidrule(lr){8-9}
Design & kg & $L$ (m) & m/s & $\hat v$ & m & $L$'s & m/s & $\hat v$ & $\eta$ \\
\midrule
Cheetah & $33$ & $0.48$ & $3.63$ & $\mathbf{1.67}$ & $0.12$ & $\mathbf{0.26}$ & $7.9$ & $\mathbf{3.65}$ & $0.20$ \\
Bear    & $110$ & $0.74$ & $\mathbf{4.33}$ & $1.61$ & $0.12$ & $0.17$ & $\mathbf{8.6}$ & $3.20$ & $\mathbf{0.47}$ \\
Giraffe & $91$ & $1.30$ & $2.17$ & $0.61$ & $\mathbf{0.20}$ & $0.15$ & $5.3$ & $1.48$ & $0.41$ \\
\bottomrule
\end{tabular}
\end{table}

\subsection{Results}


Each design converges to different capabilities, and none dominate as seen in Fig.~\ref{fig:capability} and Tab.~\ref{tab:capability}. We evaluate a robot's speed, terrain traversal, robustness, and efficiency. Speed is measured by the maximum base velocity, terrain by stair step height, and robustness with a maximum push. Drivetrain efficiency is measured as,
\begin{equation}
    \eta = \overline{P}_{\mathrm{m}} /
    (\overline{P}_{\mathrm{m}} + \overline{P}_{\mathrm{J}}),
    \label{eq:eta}
\end{equation}

where $P_{\mathrm{m}} = \sum_j |\tau_j \dot q_j|$ is the mechanical output power and $P_{\mathrm{J}} = \sum_j \tau_j^2\,\omega_{\text{NL},j}/\tau_{\text{peak},j}$ is the theoretical copper loss. In absolute units the bear is fastest and rejects the largest kick while the giraffe clears the tallest step. The efficiency of the giraffe wins at low speeds, but the bear surpasses it around 2 m/s and continues climbing at higher velocities. The cheetah wins nothing and has sizably worse efficiency compared to the others. Although, the cheetah trails just behind the bear in top speed and push stability with a much smaller frame.

\subsubsection{Normalizing by Scale}

Divide scale out and the cheetah jumps in every column of Tab.~\ref{tab:capability} other than efficiency which is not impacted by length. The adjusted speed of the cheetah barely beats out the bear's while the giraffe's relative performance drops. The adjusted terrain score flips the ordering completely. The giraffe and bear climb $15$ to $17\%$ of their leg length while the cheetah wins at $26\%$. Kick stability also flips favoring the cheetah when scaling for leg length. Altering the capability metrics for robot scale shifts the story in the cheetah's favor and provides an advantage against the poor efficiency of the QDD actuators.

\subsubsection{Performance Limiters}
We investigate what holds a design back for each task in Tab.~\ref{tab:results} which reports
the joint nearest its limit in each case. Each is stated as the $95$th percentile share of its
torque and speed ratings as a percentage. We also include the sum of the torque and speed utilization $\tau+\omega$ to represent how close the actuator is to the torque speed--curve boundary. Values above 1.00 come from when the motors are back driven which pushes an actuator past its velocity envelope.

\begin{table}[t]
\centering
\caption{Utilization at the limiting joint, knee (K) or hip pitch (H);
bold is actuator limited.}
\label{tab:results}
\small
\setlength{\tabcolsep}{1.8pt}
\begin{tabular}{@{}l lrrr@{\hspace{6pt}}lrrr@{\hspace{6pt}}lrrr@{}}
\toprule
& \multicolumn{4}{c}{Sprint} & \multicolumn{4}{c}{Kick}
& \multicolumn{4}{c}{Terrain} \\
\cmidrule(lr){2-5}\cmidrule(lr){6-9}\cmidrule(lr){10-13}
Design & & $\tau$\,\% & $\omega$\,\% & $\tau{+}\omega$
& & $\tau$\,\% & $\omega$\,\% & $\tau{+}\omega$
& & $\tau$\,\% & $\omega$\,\% & $\tau{+}\omega$ \\
\midrule
Cheetah & \textbf{K} & $\mathbf{95}$ & $\mathbf{25}$ & $\mathbf{1.05}$ & H & $69$ & $17$ & $0.60$ & K & $43$ & $13$ & $0.44$ \\
Bear    & \textbf{K} & $\mathbf{77}$ & $\mathbf{85}$ & $\mathbf{1.04}$ & \textbf{H} & $\mathbf{66}$ & $\mathbf{59}$ & $\mathbf{0.93}$ & K & $43$ & $50$ & $0.55$ \\
Giraffe & \textbf{H} & $\mathbf{96}$ & $\mathbf{105}$ & $\mathbf{1.08}$ & \textbf{H} & $\mathbf{94}$ & $\mathbf{99}$ & $\mathbf{1.06}$ & H & $76$ & $71$ & $1.00$ \\
\bottomrule
\end{tabular}
\end{table}

The velocity task is robot limited for all three instances. The cheetah's knee is torque but not speed limited while the giraffe's hip pitch motor is limited by both. The bear differs in that it is not near either the peak torque or speed, but lives at the edge of the torque--speed curve. Saturation does not make a policy optimal, but it does show the speed reached under the learned gait is restricted by the robot rather than by training. 

Push stability is actuator limited for the giraffe and bear, but not the cheetah. The smaller scale of the cheetah could explain because its dynamics may be too fast for a 50 Hz policy to recover from pushes.

The terrain curriculum consistently shows the lowest actuator utilization despite an aggressive curriculum. This is because of the termination conditions for the training environment. A robot that stays standing and does not fall gains higher reward over one that tries to climb and fails. Therefore, the limits that Fig.~\ref{fig:capability} indicate are the policy's refusal to climb because of learned robot capability from experience.

\subsubsection{Seed Variance}

Every number from Tab.~\ref{tab:capability} is a median over ten seeds, each of those a median over $256$
robot roll-outs. Reseeding moves top speed and kick by up to about a quarter of their
median and efficiency by a few percent, yet the trends survive indicating environment robustness.




\subsubsection{Sensitivity to Inertia Error}

The twin study of Sec.~\ref{sec:twins} finds that the dynamic properties of generated robots differ measurably
from those in vendor descriptions. We test this error by re-evaluating each robot using adjusted link masses which replicate the Go2 and B2 vendor-to-twin residuals. The policies are not re-trained such that we are testing the validity of our generation and evaluation pipeline, not the training environment. The results of this ablation are represented by the dashed lines in Fig.~\ref{fig:capability}. No capability trend changes with this adjustment, indicating that the evaluation pipeline is robust to the generators provided inaccuracy.
\section{Conclusion and Future Work}
\label{sec:conclusion}

We presented Draft, a parametric generation tool that compiles a short
declarative specification into a simulation-ready MJCF model. 
The generator holds no body plan, so all robots share the same compiler, leaving no bounds on the robots that can be made. Free parameters are fitted to a survey of $114$ actuators and $49$ published robot
descriptions and are incorporated into the pipeline through a design filter. A generated design is therefore not merely self-consistent but also physically credible.

Three directions would extend the work. The first is a co-design loop built on the generation and evaluation pipeline presented here, using its model fidelity. Our version of the tool keeps the designer in the loop, but a high-level optimizer could explore regions a person might not. The second is the addition of other robot subsystems, such as a dexterous hand, which would need a more extensive actuator catalog. The third is a denser sweep of the legged design space carried through to humanoids. This would turn the tradeoffs reported on just three quadrupeds into a surface that identifies optima. Each builds on what the tool already provides: a robot model that is plausible to build and easy to evaluate.

The code and a supplementary video are available at \\
\url{github.com/davidhnguyen2000/draft} and \\
\url{youtu.be/egzlEqfwwLU}.

\bibliographystyle{IEEEtran}
\bibliography{references}

\end{document}